\documentclass[conference]{IEEEtran}
\IEEEoverridecommandlockouts
\usepackage{subcaption}
\usepackage{float}
\usepackage{multirow}
\usepackage{url}
\usepackage{cite}
\usepackage{amsmath,amssymb,amsfonts}
\usepackage{algorithmic}
\usepackage{graphicx}
\usepackage{textcomp}
\usepackage{xcolor}

\newcommand{\affmark}[1]{\textsuperscript{#1}}

\usepackage{fancyhdr}
\fancypagestyle{firststyle}
{
	\fancyhf{}
    \fancyhead[C]{\fontsize{10}{10} \selectfont \textit{2025 Mexican International Conference on Computer Science (ENC) \& IEEE ICEV, Orizaba, Veracruz, México} }
    \fancyfoot[L]{\footnotesize © 2025 IEEE. Personal use of this material is permitted. Permission from IEEE must be obtained for all other uses, in any current or future media, including reprinting/republishing this material for advertising or promotional purposes, creating new collective works, for resale or redistribution to servers or lists, or reuse of any copyrighted component of this work in other works.}
}

\def\BibTeX{{\rm B\kern-.05em{\sc i\kern-.025em b}\kern-.08em
    T\kern-.1667em\lower.7ex\hbox{E}\kern-.125emX}}
\begin{document}

\title{Scalable Construction of a Lung Cancer Knowledge Base: Profiling Semantic Reasoning in LLMs\\
}

\author{%
    \IEEEauthorblockN{%
        Cesar Felipe Martínez Cisneros\affmark{a}, 
        Jesús Ulises Quiroz Bautista\affmark{a}, 
        Claudia Anahí
        Guzmán Solano\affmark{a}  
    }%
    \newline
    
    \IEEEauthorblockN{%
        Bogdan Kaleb García Rivera\affmark{a}, 
        Iván García Pacheco\affmark{a} 
    }%
    
    \newline 
    
    \IEEEauthorblockN{%
        Yalbi Itzel Balderas Martínez\affmark{b}, Kolawole John Adebayo\affmark{c}, 
        Ignacio Arroyo Fernández\affmark{a} 
    }%


  \IEEEauthorblockA{%
    \textit{\affmark{a} Universidad Tecnológica de la Mixteca}\\
     \textit{\affmark{b} Instituto Nacional de Enfermedades Respiratorias (INER) Ismael Cosío Villegas}\\
     \textit{\affmark{c} Maynooth University}\\
     \affmark{a} Huajuapan de León, Oaxaca, México,
     \affmark{b} Ciudad de México, México, \affmark{c} Maynooth, Ireland\\
    cesarmtzcisne@gmail.com, qubj980830@gs.utm.mx, claudia.guzmansolano@gmail.com, bogdanrivera@gmail.com, 
    \\ 
    iaf@gs.utm.mx, ivan@gs.utm.mx, yalbibalderas@gmail.com,  kolawole.adebayo@mu.ie  
  }%
}

\IEEEpubid{\begin{minipage}{\textwidth}\\[12pt]
    \centering
    \footnotesize
    © 2025 IEEE. Personal use of this material is permitted. Permission from IEEE must be obtained for all other uses, in any current or future media, including reprinting/republishing this material for advertising or promotional purposes, creating new collective works, for resale or redistribution to servers or lists, or reuse of any copyrighted component of this work in other works.
\end{minipage}}

\maketitle
\thispagestyle{firststyle}
\begin{abstract}
The integration of Large Language Models (LLMs) into biomedical research offers new opportunities for domain-specific reasoning and knowledge representation. However, their performance depends heavily on the semantic quality of training data. In oncology, where precision and interpretability are vital, scalable methods for constructing structured knowledge bases are essential for effective fine-tuning. This study presents a pipeline for developing a lung cancer knowledge base using Open Information Extraction (OpenIE). The process includes: (1) identifying medical concepts with the MeSH thesaurus; (2) filtering open-access PubMed literature with permissive licenses (CC0); (3) extracting (subject, relation, object) triplets using OpenIE method; and (4) enriching triplet sets with Named Entity Recognition (NER) to ensure biomedical relevance. The resulting triplet sets provide a domain-specific, large-scale, and noise-aware resource for fine-tuning LLMs. We evaluated T5 models fine-tuned on this dataset through Supervised Semantic Fine-Tuning. Comparative assessments with  ROUGE and BERTScore show significantly improved performance and semantic coherence, demonstrating the potential of OpenIE-derived resources as scalable, low-cost solutions for enhancing biomedical NLP.
\end{abstract}


\section{Introduction}
The rapid expansion of scientific knowledge, particularly in the biomedical domain, has created an urgent need for tools that facilitate efficient access to and understanding of large-scale information \cite{harper2018agbiodata}. Large Language Models (LLMs) have transformed Natural Language Processing (NLP) \cite{min2023recent}, providing new opportunities for healthcare professionals and researchers to engage with specialized literature.

Despite these advances, LLMs face critical challenges related to input data quality. A major limitation is the absence of high-coverage, accessible knowledge resources, which constrains knowledge-based Artificial Intelligence (AI) systems. Automated data collection methods reduce time and cost but often propagate errors. Scaling model capacity mitigates some noise, yet this strategy remains limited to applications tolerant of residual inaccuracies. Manual curation ensures high-quality data but is prohibitively expensive and time-consuming, resulting in datasets with restricted coverage.

Knowledge Graphs and Ontologies exemplify curated resources that can reduce hallucinations in LLMs. However, they reveal a trade-off across scalability, noise, flexibility, and precision. While precise resources constrain hallucinations, they fail to support broader applications. Scalable resources, in contrast, enable wider coverage but inevitably contain noise, demanding additional quality control. This paradox particularly challenges public institutions and non-profits, which must maximize AI utility under resource constraints.

Traditional approaches for knowledge base construction, such as relation extraction pipelines and curated graphs, also face limitations: the former are inflexible and annotation-dependent, while the latter are costly \cite{jehangir2023survey,xu2013open}. Open Information Extraction (OpenIE) offers an alternative by providing automated, low-cost, and scalable knowledge acquisition \cite{romero2023mapping}. Unlike schema-dependent methods, OpenIE processes large volumes of text with extensive coverage, facilitating scalable knowledge generation.

LLMs exhibit strong reasoning abilities, though the mechanisms remain unclear. A prevailing theory suggests these abilities rely heavily on semantic structures implicitly learned from training data \cite{tang2023large}. Consequently, lightweight, cost-effective pipelines are needed to build large-scale knowledge bases, support small language models with robust reasoning, and provide affordable means to evaluate semantic coherence.

This research addresses these challenges by proposing a pipeline to construct a lightweight, domain-specific knowledge base on lung cancer for semantic supervision. The pipeline employs OpenIE to generate triplets and integrates filtering mechanisms to ensure biomedical relevance. To evaluate its impact, we analyze the semantic reasoning abilities of LLMs fine-tuned with this knowledge base, specifically assessing their capacity to generate accurate object phrases from subject–relation pairs, simulating structured inference in biomedical contexts. We evaluated T5 models fine-tuned on this dataset through Supervised Semantic Fine-Tuning. Comparative assessments with ROUGE and BERTScore show significantly improved performance and semantic coherence, demonstrating the potential of OpenIE-derived resources as scalable, low-cost solutions for enhancing biomedical reasoning.

\section{Related Literature}

Prior work has focused on building medical knowledge graphs using pattern-mining \cite{shi2017semantic}, electronic health records \cite{li2020real}, and tensor factorization for comorbidity prediction \cite{biswas2019relation}. Other approaches utilize pipelines with tools like SciSpacy and BioBERT for specific tasks like protein-protein interaction extraction \cite{parmar2020biomedical} or employ hybrid neural networks for relation extraction \cite{wu2021bioie}. Our work is similar to \cite{balderas2025semantic} in its use of smaller models but differentiates itself by focusing on a scalable OpenIE pipeline to generate the knowledge base itself.

\section{Theoretical Background}

The rapid evolution of AI and NLP has made it possible to process large amounts of information, and the quality of that information is crucial for working with LLMs (\cite{devlin2019bert}, \cite{rejeleene2024towards}). Knowledge resources, such as lexical databases like WordNet \cite{miller1995wordnet} and specialized biomedical ontologies like MeSH \cite{fb1963medical}, along with knowledge graphs like BioKG \cite{walsh2020biokg}, are essential for fine-tuning LLMs by providing high-quality data that reduces hallucinations and improves semantic accuracy \cite{baldazzi2023fine}. Open Information Extraction (OpenIE) is a flexible and efficient alternative to conventional methods, as it does not require predefined schemas, offering greater recall and expressiveness in triplet extraction \cite{xu2013open,sheikhalishahi2019natural}. While it may produce noisy triplets, OpenIE is more economical and lowers barriers to research \cite{romero2023mapping,ji2021survey,mesquita2013effectiveness}. This method is crucial for developing robust semantic reasoning, which involves inferring missing elements in subject-predicate-object triplets \cite{fang2021benchmarking, luong2015effective, vaswani2017attention}. This process, known as a Knowledge Base Completion (KBC) problem, is modeled using neural networks that learn distributed phrase representations to predict an object given a subject-predicate pair \cite{fang2021benchmarking, luong2015effective, vaswani2017attention}.

\section{Methodology and Implementation}

\subsection{Methodology}

The project's methodology was divided into two phases: knowledge base acquisition and application and use of the dataset.

\begin{itemize}
\item \textbf{Knowledge base acquisition}

\begin{enumerate}
    \item \textbf{Corpus Acquisition:} A corpus of approximately 2,901,198 biomedical articles (190 GB) was downloaded from the PubMed Central (PMC).
    
    \item \textbf{Terminology and Filtering:} Identified 5 key MeSH terms for 'lung cancer.' Filtered the PMC corpus using a k-NN search, yielding 37,951 relevant articles.

    \item \textbf{License Verification:} Used the NCBI API to filter for articles with permissive license (CC0), resulting in a final set of 2,967 articles.
    
    \item \textbf{Triplet Extraction:} Processed the licensed articles with the Stanford CoreNLP OpenIE module to extract 7.5 million (Subject, Relation, Object) triplets from CC0-licensed articles.
    \item \textbf{Biomedical Entity Filtering:} Employed a "zero-shot" Hugging Face model for Named Entity Recognition (NER) to score triplets. We retained only those where both the subject and object had a $>$80\% probability of being a biomedical entity, reducing the dataset to ~13k high-quality triplets.
\end{enumerate}

\item \textbf{Application and use of the dataset}

To validate the dataset, we conducted a comprehensive evaluation of two T5 model variants: a base T5 model and a PubMed fine-tuned model with LoRa layers trained on PubMed abstract summarization datasets. The evaluation framework comprised three main methodological components:
\begin{enumerate}

    \item \textbf{Data preprocessing:} We created both filtered (biomedical entity probability    
    $>$80\%) and unfiltered datasets, followed by randomization and partitioning into training (10,000 triplets), testing (1,000 triplets), and validation (200 triplets) sets stored in CSV format.

    \item \textbf{Model fine-tuning and evaluation:} We implemented fine-tuning using the Trainer class with respective tokenizers (T5Tokenizer for T5-base; T5 Tokenizer Fast for T5-PubMed). Both models were evaluated in zero-shot inference and post-fine-tuning scenarios using BERTScore (precision, recall, F1-score) and ROUGE metrics on the validation set.

    \item \textbf{Results analysis:} We performed comparative evaluation through statistical tables and visualization plots (histograms and KDE) to assess performance differences and the impact of randomness on F1-score values across filtered and unfiltered conditions.

\end{enumerate}

\end{itemize}
All experiments were conducted on commodity hardware featuring dual RTX-4090 GPUs with 24GB VRAM and 32GB CPU RAM.\footnote{Model details: T5-Base \url{https://huggingface.co/google-t5/t5-base}; T5-PubMed \url{https://huggingface.co/Kevincp560/t5-base-finetuned-pubmed}}

For the training configuration, a maximum sequence length of 50 tokens was utilized. Models were trained for a total of 4 epochs employing the Adafactor optimizer. A per-device batch size of 50 was applied for both training and evaluation. The evaluation, checkpoint saving, and logging strategies were uniformly set to ``epoch''. Experiment metrics were tracked using Weights \& Biases. To ensure optimal performance, the best model was identified based on the lowest evaluation loss and automatically loaded upon completion of training.

\section{Results and Discussion}

\noindent
The pipeline filtering stages progressed from an initial acquisition of 2.9 million articles down to a final result of 13,000 lung cancer knowledge triplets, as shown in Table \ref{tab:triplet_pipeline}. 

The corpus acquisition stage gathered comprehensive biomedical literature, followed by terminology-based filtering that reduced the dataset to 37.9K relevant articles. License verification ensured compliance with usage rights, resulting in 2.9K articles suitable for processing. The triplet extraction phase generated 7.5M candidate triplets from these articles. Finally, biomedical entity filtering refined the results to 13K high-quality knowledge triplets from 1.4K articles.

\begin{table}[htbp]
\centering
\caption{Summary of the Knowledge Triplet Extraction Process}
\begin{tabular}{|l|c|c|}
\hline
\textbf{Pipeline Stage}       & \textbf{Articles} & \textbf{Triplets} \\
\hline
Corpus Acquisition            & 2.9M   & --     \\
Terminology and Filtering     & 37.9K  & --     \\
License Verification          & 2.9K   & --     \\
Triplet Extraction            & 2.9K   & 7.5M  \\
Biomedical Entity Filtering   & 1.4K    & 13K    \\
\hline
\end{tabular}
\label{tab:triplet_pipeline}
\end{table}

The results presented herein are the most representative of the 8 experiments conducted in total. Although the remaining experiments yielded plots similar to those shown in Figure 1, only the most significant findings were included for brevity. The omitted experiments exhibited behavior analogous to Figure 1a, wherein no discernible difference between the two distributions was observed. However, this changed significantly with the use of filtered data, which produced plots similar to Figure 1b, illustrating a clear divergence in results. The experiments not included in the main paper encompass the training of the T5-base model on both filtered and unfiltered data, as well as the training of the T5-pubmed model under these same two data conditions.

\subsubsection{Differential Metric Sensitivity to Semantic Fine-tuning}

The most striking finding is the differential response of evaluation metrics to semantic fine-tuning. ROUGE metrics exhibit dramatic improvements (see Table \ref{tab:rouge_results_filtrado}): ROUGE-1 F1-Score increases from 0.0305 to 0.2003 for T5-Base ($\sim657\%$ improvement) and from 0.0274 to 0.2025 for T5-PubMed ($\sim739\%$ improvement). Similarly, ROUGE-2 F1-Score shows increases of 1275\% and 1583\% respectively. In contrast, BertScore F1-Score improvements are more moderate: 21\% for T5-Base (0.6289 to 0.7606) and 19\% for T5-PubMed (0.6363 to 0.7582). See Table \ref{tab:bertscore_results_filtrado}. 
These comparisons were observed using our NER probability filtered versions of the data, which showed substantially better results compared to non-filtered data where ROUGE-1 F1-Scores reached only 0.0651 (T5-Base) and 0.0694 (T5-PubMed) after fine-tuning, and BertScore F1-Scores plateaued at 0.6892 (T5-Base) and 0.6944 (T5-PubMed) respectively under non-filtered conditions.

This differential sensitivity reflects fundamental metric differences: ROUGE measures lexical overlap, directly benefiting from structured triplet training and precise biomedical terminology learning, while BertScore captures semantic similarity through contextual embeddings, yielding higher baseline values but modest improvements. However, low ROUGE scores may not indicate poor performance, as later verified using a Meaning-Based Selectional Preference Test (MSPT) \cite{balderas2025semantic}.

\begin{table}[htbp]
\centering
\caption{ROUGE Evaluation Results of T5-base models under zero-shot inference and semantic fine-tuning \textbf{with} NER-probability filtering}
\label{tab:rouge_results_filtrado}
\begin{tabular}
{|p{1.1cm}|p{1.4cm}|p{1.0cm}|p{1.0cm}|p{1.0cm}|p{1.1cm}|}
\hline
\textbf{Metric} & \textbf{Component} & 
\textbf{T5-Base} & 
\textbf{T5-Base (Fine-tuned)} & 
\textbf{T5-PubMed} & 
\textbf{T5-PubMed (Fine-tuned)} \\
\hline
\multirow{3}{*}{ROUGE-1} & Precision & 0.1023 & 0.2077 & 0.1018 & 0.2095 \\
\cline{2-6}
& Recall & 0.0197 & 0.2227 & 0.0165 & 0.2265 \\
\cline{2-6}
& F1-Score & 0.0305 & 0.2003 & 0.0274 & 0.2025 \\
\hline
\multirow{3}{*}{ROUGE-2} & Precision & 0.0293 & 0.1003 & 0.0293 & 0.1008 \\
\cline{2-6}
& Recall & 0.0047 & 0.1019 & 0.0034 & 0.1012 \\
\cline{2-6}
& F1-Score & 0.0073 & 0.0931 & 0.0059 & 0.0934 \\
\hline
\multirow{3}{*}{ROUGE-L} & Precision & 0.1023 & 0.2025 & 0.1018 & 0.2030 \\
\cline{2-6}
& Recall & 0.0197 & 0.2181 & 0.0165 & 0.2208 \\
\cline{2-6}
& F1-Score & 0.0305 & 0.1956 & 0.0274 & 0.1967 \\
\hline
\end{tabular}    
\end{table}

\begin{table}[htbp]
\centering
\caption{BertScore Evaluation Results of T5-base models under zero-shot and semantic fine-tuning \textbf{with} filtered data}
\label{tab:bertscore_results_filtrado}
\begin{tabular}{|p{1.3cm}|p{1.1cm}|p{1.5cm}|p{1.5cm}|p{1.6cm}|}
\hline
\textbf{Component} & 
\textbf{T5-Base} & 
\textbf{T5-Base (Fine-tuned)} & 
\textbf{T5-PubMed} & 
\textbf{T5-PubMed (Fine-tuned)} \\
\hline
Precision & 0.5925 & 0.7668 & 0.6049 & 0.7629 \\
\hline
Recall & 0.6722 & 0.7562 & 0.6727 & 0.7554 \\
\hline
F1-Score & 0.6289 & 0.7606 & 0.6363 & 0.7582 \\
\hline
\end{tabular}
\end{table}

\subsubsection{Model Convergence Following Domain-Specific Fine-tuning}

A remarkable pattern emerges in the post fine-tuning performance: T5-Base and T5-PubMed models converge to nearly identical results across all metrics. For BertScore F1, the difference narrows from 0.0074 (pre-training) to 0.0024 (post fine-tuning). Similarly, ROUGE metrics show convergence, with ROUGE-1 F1 differences reducing from 0.0031 to 0.0022.
This convergence suggests that semantic fine-tuning on domain-specific triplets acts as a powerful equalizing mechanism. The structured nature of the OpenIE-derived knowledge base provides explicit biomedical concept associations that override the advantages of domain-specific pre-training, at least in the case of biomedical text summarization. This finding has important implications for practical applications: researchers may achieve comparable performance using general-purpose models with targeted semantic fine-tuning rather than investing in domain-specific pretraining, reducing computational costs and complexity.
The convergence also indicates that the triplet-based supervision provides sufficiently rich semantic information to reshape model representations toward domain-specific requirements, regardless of initial training domain biases.

\subsubsection{Precision-Recall Rebalancing in ROUGE Metrics}

The ROUGE metrics reveal a significant rebalancing between precision and recall components following fine-tuning. In the zero-shot condition, ROUGE-1 precision (0.1023 for T5-Base) substantially exceeds recall (0.0197), creating a 5.2:1 precision-to-recall ratio. Post fine-tuning, this ratio dramatically shifts to near parity: precision (0.2077) and recall (0.2227) achieve a balanced 0.93:1 ratio (see Table \ref{tab:rouge_results_filtrado}).
This rebalancing reflects a fundamental shift in model behavior. Pre-fine-tuning models generate conservative outputs with limited lexical overlap (low recall) but what they do generate tends to be relevant (relatively higher precision). The poor recall suggests models struggle to produce the specific biomedical terminology required for the task. Semantic fine-tuning addresses this limitation by teaching models to generate longer, more comprehensive outputs that better match reference texts while maintaining relevance.
The improvement in recall indicates that triplet-based training successfully expands the model's active biomedical vocabulary, enabling generation of domain-specific terms and phrases that were previously inaccessible, even with biomedical summarization fine-tuning. This is particularly crucial for biomedical applications where precise terminology and comprehensive coverage of relevant concepts are essential for clinical utility.

Both for filtered and unfiltered data, fine-tuning improves the scores compared to the zero-shot version. With unfiltered triplets, the ROUGE metrics remain low, particularly ROUGE-2, while BertScore reflects marginal gains. With filtered triplets, fine-tuning with only 10,000 examples enhances both BertScore and ROUGE performance metrics. The jump in precision and F1 is remarkable, especially for T5-Base. Undoubtedly, filtering by probability of biomedical entities improves the results. Notably, a slight superiority of the base model over T5-PubMed was observed in the ROUGE metric prior to fine-tuning. However, this trend reversed after fine-tuning, with the T5-PubMed model performing marginally better.

\begin{figure}[h]
  \centering
  \begin{minipage}{0.48\textwidth}
    \centering
    \includegraphics[width=0.95\linewidth]{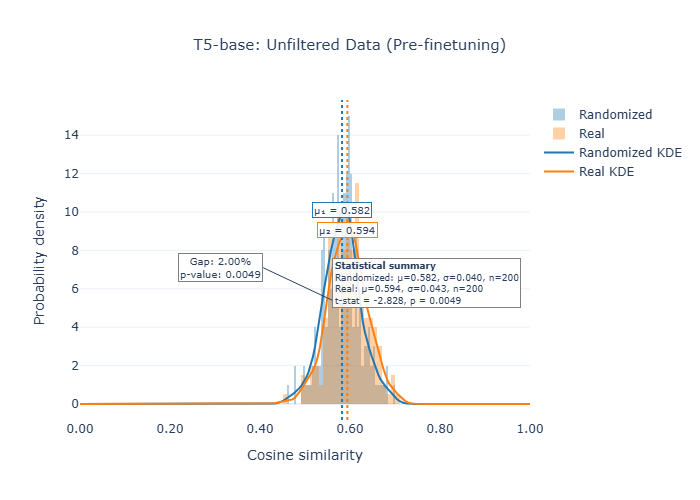}
    \subcaption{T5-Pubmed (Pre-finetuning)}
  \end{minipage}
  \hfill
  \begin{minipage}{0.48\textwidth}
    \centering
    \includegraphics[width=0.95\linewidth]{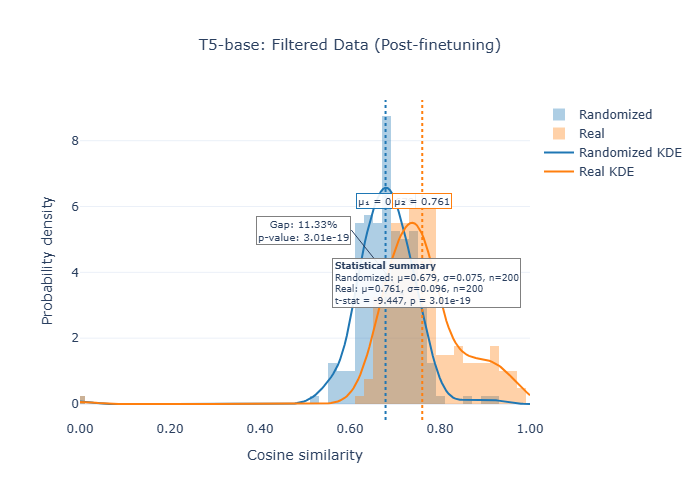}
    \subcaption{T5-Pubmed (Post-finetuning)}
  \end{minipage}
  
  \caption{
    F1-BERTscore distributions for filtered data,
    showing model performance before (left column) and after (right column) fine-tuning.
    Each subplot compares two score distributions calculated over 200 samples:
    (i) actual performance (orange curve), and (ii) random baseline (blue curve).
  }
  \label{fig:combined KDE e histograma}
\end{figure}

The improvement between using filtered and non-filtered data was not evident from purely seeing at BERTscores (relatively high values in both cases). However, by seeing at Figure \ref{fig:combined KDE e histograma} this view is completely changed. To assess the statistical significance of these improvements, we compared F1 BERTscores under two conditions: (1) model predictions evaluated against correct expected outputs, and (2) model predictions evaluated against randomized expected outputs (MSPT) (Table \ref{tab:combined_bertscore_results}). This methodology enables us to determine whether the models have acquired semantically meaningful knowledge rather than generating random meanings with inflated BERTScores.
These results are also summarized in Table \ref{tab:combined_comparison}, which additionally include the p-value and gap values (the difference between means) for tests conducted with the T5 model using filtered data and the T5-PubMed model using unfiltered data. These values were similar to those obtained with the T5-PubMed and T5 models (Figure \ref{fig:combined KDE e histograma}).

Clear patterns are identified: a) Fine-tuning consistently helps to generate relevant biomedical meanings. b) NER filtering contributes more quality than quantity; despite using a much smaller dataset (11,200 triplets), the results with filtered triplets are superior ($p<3\times10^{-19}$). c) T5-PubMed does not always clearly outperform T5-Base; the results show that the difference is small, and in some cases, fine-tuned T5-Base even matches or slightly surpasses it; however, MSPT showed the most significant (though small) improvement with Pubmed fine-tuning ($p<10^{-24}$). Regarding the low performance with unfiltered data, we think that noise and ambiguity in the unfiltered triplets likely hinder learning. Filtered data with NER probability performs better because the filtering removed noise and redundancies, leaving more reliable noun phrases. The small differences between T5-Base and T5-PubMed may be due to the task (object prediction in triplets) being very different from T5-PubMed's biomedical pre-training. Finally, the contrast between BertScore and ROUGE is explained by the former capturing phrase meaning, while the latter depends on exact token matches.

\begin{table}[h]
\centering
\caption{BERTScore Analysis of T5-base and T5-PubMed: Zero-Shot versus Fine-Tuned Performance with Filtered Randomized Gold Labels}
\begin{tabular}{|p{1.2cm}|p{1.1cm}|p{1.5cm}|p{1.5cm}|p{1.5cm}|}
\hline
& \textbf{T5-Base} & \textbf{T5-Base (Fine-tuned)} & \textbf{T5-PubMed} & \textbf{T5-PubMed (Fine-tuned)} \\
\hline
\textbf{Metric} & Filtered & Filtered & Filtered & Filtered \\
\hline
Precision & 0.5726 & 0.6838 & 0.5826 & 0.6808 \\
Recall & 0.6400 & 0.6757 & 0.6409 & 0.6744 \\
F1-Score & 0.6039 & 0.6790 & 0.6099 & 0.6768 \\
\hline
\textbf{Metric} & Unfiltered & Unfiltered & Unfiltered & Unfiltered \\
\hline
Precision & 0.5431 & 0.6570 & 0.5477 & 0.6621 \\
Recall & 0.6296 & 0.6545 & 0.6312 & 0.6577 \\
F1-Score & 0.5824 & 0.6546 & 0.5857 & 0.6588 \\
\hline
\end{tabular}
\label{tab:combined_bertscore_results}
\end{table}

\begin{table}[H]
\centering
\caption{Comparison of models with filtered and unfiltered data pre and post fine-tuning}
\begin{tabular}{|l|c|c|c|c|}
\hline
& \multicolumn{2}{c|}{\textbf{T5-base}} & \multicolumn{2}{c|}{\textbf{T5-Pubmed}} \\
\cline{2-5}
\textbf{Metric} & \textbf{Unfiltered} & \textbf{Filtered} & \textbf{Unfiltered} & \textbf{Filtered} \\
\hline
p-value (pre-FT) & 0.0049 & 5.49e-08 & 0.0085 & 1.18e-08 \\
p-value (post-FT) & 0.0016 & 3.01e-19 & 2.45e-06 & 1.14e-24 \\
\hline
p-value difference & 0.0033 & 5.49e-08 & 0.00849755 & 1.18e-08 \\
\hline
Gap pre-FT (\%) & 2.00 & 4.06 & 1.96 & 4.24 \\
Gap post-FT (\%) & 5.15 & 11.33 & 5.27 & 11.34 \\
\hline
Gap Increase (\%) & 3.15 & 7.27 & 3.31 & 7.10 \\
\hline
\end{tabular}
\label{tab:combined_comparison}
\end{table}

\section{Conclusions}

 
The differential metric responses highlight mechanisms behind semantic fine-tuning effectiveness. Substantial ROUGE gains show that OpenIE-derived triplets effectively teach lexical and syntactic patterns in biomedical discourse, reflecting their structured ability to encode relations between entities using natural language. More moderate BertScore improvements indicate that while semantic coherence improves, contextual similarity gains are less pronounced. This suggests that deeper semantic understanding in specialized domains may require additional supervision beyond triplet-based learning, including discourse-level coherence or causal reasoning. The convergence findings also carry practical implications: biomedical NLP projects may achieve competitive performance by combining general-purpose models with targeted semantic fine-tuning, optimizing resource allocation without always requiring domain-specific pretraining.

For future work, we plan to assess the generalizability of our approach by applying it to corpora related to other diseases beyond lung cancer. We also intend to explore and compare more advanced Open Information Extraction (Open IE) methods. The baseline system used in this study may have performance limitations; thus, investigating state-of-the-art Open IE techniques could significantly enhance triplet quality. In conjunction with this, integrating methods to estimate triplet factuality is a crucial step for improving the veracity of the extracted information. Furthermore, expanding the training corpus using permissively licensed data sources would likely yield significant performance gains by providing a larger volume of training examples.

Subsequent experiments will involve a multi-stage training strategy, entailing pre-training on large-scale, general-purpose datasets (such as ConceptNet or SNLI) that address analogous tasks, prior to fine-tuning on our domain-specific corpus. Finally, we plan to conduct a comprehensive analysis comparing different model architectures and sizes, as well as varying dataset scales, to thoroughly evaluate the robustness and scalability of our fine-tuning methodology.

\section{Acknowledgments}
We express our deepest and most sincere gratitude to the Secretariat of Science, Humanities, Technology and Innovation (SECIHTI) for the trust placed in us and the valuable funding granted to our project with key code CF-2023-I-2854. This crucial support not only ensures the viability of our research but also drives the advancement of the frontier of knowledge in our field, allowing us to work on solutions that aspire to have a significant impact.

\bibliographystyle{IEEEtran}
\bibliography{referencias}

\begin{thebibliography}{10}
\providecommand{\url}[1]{#1}
\csname url@samestyle\endcsname
\providecommand{\newblock}{\relax}
\providecommand{\bibinfo}[2]{#2}
\providecommand{\BIBentrySTDinterwordspacing}{\spaceskip=0pt\relax}
\providecommand{\BIBentryALTinterwordstretchfactor}{4}
\providecommand{\BIBentryALTinterwordspacing}{\spaceskip=\fontdimen2\font plus
\BIBentryALTinterwordstretchfactor\fontdimen3\font minus \fontdimen4\font\relax}
\providecommand{\BIBforeignlanguage}[2]{{%
\expandafter\ifx\csname l@#1\endcsname\relax
\typeout{** WARNING: IEEEtran.bst: No hyphenation pattern has been}%
\typeout{** loaded for the language `#1'. Using the pattern for}%
\typeout{** the default language instead.}%
\else
\language=\csname l@#1\endcsname
\fi
#2}}
\providecommand{\BIBdecl}{\relax}
\BIBdecl

\bibitem{harper2018agbiodata}
L.~Harper, J.~Campbell, E.~K. Cannon, S.~Jung, M.~Poelchau, R.~Walls, C.~Andorf, E.~Arnaud, T.~Z. Berardini, C.~Birkett \emph{et~al.}, ``Agbiodata consortium recommendations for sustainable genomics and genetics databases for agriculture,'' \emph{Database}, vol. 2018, p. bay088, 2018.

\bibitem{min2023recent}
B.~Min, H.~Ross, E.~Sulem, A.~P.~B. Veyseh, T.~H. Nguyen, O.~Sainz, E.~Agirre, I.~Heintz, and D.~Roth, ``Recent advances in natural language processing via large pre-trained language models: A survey,'' \emph{ACM Computing Surveys}, vol.~56, no.~2, pp. 1--40, 2023.

\bibitem{jehangir2023survey}
B.~Jehangir, S.~Radhakrishnan, and R.~Agarwal, ``A survey on named entity recognition—datasets, tools, and methodologies,'' \emph{Natural Language Processing Journal}, vol.~3, p. 100017, 2023.

\bibitem{xu2013open}
Y.~Xu, M.-Y. Kim, K.~M. Quinn, R.~Goebel, and D.~Barbosa, ``Open information extraction with tree kernels,'' in \emph{Proceedings of the 2013 Conference of the North American Chapter of the Association for Computational Linguistics: Human Language Technologies}, 2013, pp. 868--877.

\bibitem{romero2023mapping}
J.~Romero and S.~Razniewski, ``Mapping and cleaning open commonsense knowledge bases with generative translation,'' in \emph{International Semantic Web Conference}.\hskip 1em plus 0.5em minus 0.4em\relax Springer, 2023, pp. 368--387.

\bibitem{tang2023large}
X.~Tang, Z.~Zheng, J.~Li, F.~Meng, S.-C. Zhu, Y.~Liang, and M.~Zhang, ``Large language models are in-context semantic reasoners rather than symbolic reasoners,'' \emph{arXiv preprint arXiv:2305.14825}, 2023.

\bibitem{shi2017semantic}
L.~Shi, S.~Li, X.~Yang, J.~Qi, G.~Pan, and B.~Zhou, ``Semantic health knowledge graph: semantic integration of heterogeneous medical knowledge and services,'' \emph{BioMed research international}, vol. 2017, no.~1, p. 2858423, 2017.

\bibitem{li2020real}
L.~Li, P.~Wang, J.~Yan, Y.~Wang, S.~Li, J.~Jiang, Z.~Sun, B.~Tang, T.-H. Chang, S.~Wang \emph{et~al.}, ``Real-world data medical knowledge graph: construction and applications,'' \emph{Artificial intelligence in medicine}, vol. 103, p. 101817, 2020.

\bibitem{biswas2019relation}
S.~Biswas, P.~Mitra, and K.~S. Rao, ``Relation prediction of co-morbid diseases using knowledge graph completion,'' \emph{IEEE/ACM Transactions on Computational Biology and Bioinformatics}, vol.~18, no.~2, pp. 708--717, 2019.

\bibitem{parmar2020biomedical}
J.~Parmar, W.~Koehler, M.~Bringmann, K.~S. Volz, and B.~Kapicioglu, ``Biomedical information extraction for disease gene prioritization,'' \emph{arXiv preprint arXiv:2011.05188}, 2020.

\bibitem{wu2021bioie}
J.~Wu, R.~Zhang, T.~Gong, Y.~Liu, C.~Wang, and C.~Li, ``Bioie: Biomedical information extraction with multi-head attention enhanced graph convolutional network,'' in \emph{2021 IEEE International Conference on Bioinformatics and Biomedicine (BIBM)}.\hskip 1em plus 0.5em minus 0.4em\relax IEEE, 2021, pp. 2080--2087.

\bibitem{balderas2025semantic}
Y.~I. Balderas-Mart{\'\i}nez, J.~A. S{\'a}nchez-Rojas, A.~T{\'e}llez-Vel{\'a}zquez, F.~Ju{\'a}rez~Mart{\'\i}nez, R.~Cruz-Barbosa, E.~Guzm{\'a}n-Ram{\'\i}rez, I.~Garc{\'\i}a-Pacheco, and I.~Arroyo-Fern{\'a}ndez, ``Semantic reasoning using standard attention-based models: An application to chronic disease literature,'' \emph{Big Data and Cognitive Computing}, vol.~9, no.~6, p. 162, 2025.

\bibitem{devlin2019bert}
J.~Devlin, M.-W. Chang, K.~Lee, and K.~Toutanova, ``Bert: Pre-training of deep bidirectional transformers for language understanding,'' in \emph{Proceedings of the 2019 conference of the North American chapter of the association for computational linguistics: human language technologies, volume 1 (long and short papers)}, 2019, pp. 4171--4186.

\bibitem{rejeleene2024towards}
R.~Rejeleene, X.~Xu, and J.~Talburt, ``Towards trustable language models: Investigating information quality of large language models,'' \emph{arXiv preprint arXiv:2401.13086}, 2024.

\bibitem{miller1995wordnet}
G.~A. Miller, ``Wordnet: a lexical database for english,'' \emph{Communications of the ACM}, vol.~38, no.~11, pp. 39--41, 1995.

\bibitem{fb1963medical}
R.~FB, ``Medical subject headings.'' \emph{Bulletin of the Medical Library Association}, vol.~51, pp. 114--116, 1963.

\bibitem{walsh2020biokg}
B.~Walsh, S.~K. Mohamed, and V.~Nov{\'a}{\v{c}}ek, ``Biokg: A knowledge graph for relational learning on biological data,'' in \emph{Proceedings of the 29th ACM International Conference on Information \& Knowledge Management}, 2020, pp. 3173--3180.

\bibitem{baldazzi2023fine}
T.~Baldazzi, L.~Bellomarini, S.~Ceri, A.~Colombo, A.~Gentili, and E.~Sallinger, ``Fine-tuning large enterprise language models via ontological reasoning,'' in \emph{International Joint Conference on Rules and Reasoning}.\hskip 1em plus 0.5em minus 0.4em\relax Springer, 2023, pp. 86--94.

\bibitem{sheikhalishahi2019natural}
S.~Sheikhalishahi, R.~Miotto, J.~T. Dudley, A.~Lavelli, F.~Rinaldi, and V.~Osmani, ``Natural language processing of clinical notes on chronic diseases: systematic review,'' \emph{JMIR medical informatics}, vol.~7, no.~2, p. e12239, 2019.

\bibitem{ji2021survey}
S.~Ji, S.~Pan, E.~Cambria, P.~Marttinen, and P.~S. Yu, ``A survey on knowledge graphs: Representation, acquisition, and applications,'' \emph{IEEE transactions on neural networks and learning systems}, vol.~33, no.~2, pp. 494--514, 2021.

\bibitem{mesquita2013effectiveness}
F.~Mesquita, J.~Schmidek, and D.~Barbosa, ``Effectiveness and efficiency of open relation extraction,'' in \emph{Proceedings of the 2013 Conference on Empirical Methods in Natural Language Processing}, 2013, pp. 447--457.

\bibitem{fang2021benchmarking}
T.~Fang, W.~Wang, S.~Choi, S.~Hao, H.~Zhang, Y.~Song, and B.~He, ``Benchmarking commonsense knowledge base population with an effective evaluation dataset,'' \emph{arXiv preprint arXiv:2109.07679}, 2021.

\bibitem{luong2015effective}
M.-T. Luong, H.~Pham, and C.~D. Manning, ``Effective approaches to attention-based neural machine translation,'' \emph{arXiv preprint arXiv:1508.04025}, 2015.

\bibitem{vaswani2017attention}
A.~Vaswani, N.~Shazeer, N.~Parmar, J.~Uszkoreit, L.~Jones, A.~N. Gomez, {\L}.~Kaiser, and I.~Polosukhin, ``Attention is all you need,'' \emph{Advances in neural information processing systems}, vol.~30, 2017.

\end{thebibliography}


\vspace{12pt}

\end{document}